\documentclass{article}

\usepackage{microtype}
\usepackage{graphicx}
\usepackage{subfigure}
\usepackage{booktabs} 

\usepackage{hyperref}

\usepackage[accepted]{icml2024}

\usepackage{amsmath}
\usepackage{amssymb}
\usepackage{mathtools}
\usepackage{amsthm}

\usepackage[capitalize,noabbrev]{cleveref}

\theoremstyle{plain}
\newtheorem{theorem}{Theorem}[section]
\newtheorem{proposition}[theorem]{Proposition}
\newtheorem{lemma}[theorem]{Lemma}
\newtheorem{corollary}[theorem]{Corollary}
\theoremstyle{definition}

\theoremstyle{remark}
\newtheorem{remark}[theorem]{Remark}

\usepackage[disable,textsize=tiny]{todonotes}

\theoremstyle{definition}
\newtheorem{example}[theorem]{Example}

\newcommand{\R}{\mathbb{R}}
\newcommand{\N}{\mathbb{N}}

\newcommand{\norm}[1]{\left\lVert #1\right\rVert}

\DeclareMathOperator*{\argmin}{argmin}
\DeclareMathOperator*{\argmax}{argmax}

\DeclareMathOperator{\tr}{Tr}

\DeclareMathOperator{\Proj}{Proj}

\DeclareMathOperator{\vecto}{vec}
\DeclareMathOperator{\image}{Im}

\begin{document}

\twocolumn[
\icmltitle{Multivariate Online Linear Regression for Hierarchical Forecasting}




\begin{icmlauthorlist}
\icmlauthor{Massil Hihat}{lopf,lpsm}
\icmlauthor{Guillaume Garrigos}{lopf,lpsm}
\icmlauthor{Adeline Fermanian}{lopf}
\icmlauthor{Simon Bussy}{lopf}
\end{icmlauthorlist}

\icmlaffiliation{lopf}{LOPF, Califrais’ Machine Learning Lab, Paris, France}
\icmlaffiliation{lpsm}{LPSM, UMR 8001, CNRS, Sorbonne University, Paris Cité University, Paris, France}

\icmlcorrespondingauthor{Massil Hihat}{hihat@lpsm.paris}



\vskip 0.3in
]



\printAffiliationsAndNotice{}  

\begin{abstract}
    In this paper, we consider a deterministic online linear regression model where we allow the responses to be multivariate. To address this problem, we introduce MultiVAW, a method that extends the well-known Vovk-Azoury-Warmuth algorithm to the multivariate setting, and show that it also enjoys logarithmic regret in time. We apply our results to the online hierarchical forecasting problem and recover an algorithm from this literature as a special case, allowing us to relax the hypotheses usually made for its analysis.
\end{abstract}

\section{Introduction}

In classical linear regression, a learner is given a fixed set of pairs $(x_i, y_i)_{i=1, \dots, n}$ from which a parameter $\theta$ is learned such that $y_i \approx \theta^\top x_i$. In many applications, however, the pairs $(x_i,y_i)$ are revealed only sequentially: as time passes, nature reveals new examples. This is the framework of online linear regression: at each time step, we assume that a learner receives some features $x_t$, updates its current parameter $\theta_t$, outputs a prediction of the form $\hat{y}_t =  \theta_t^\top x_t$ and then the true value $y_t$ is revealed. This setting is common, for example, in finance \citep{zhang2022application} or energy load forecasting \citep{ohf_bregere_2022}. It has also applications in model-based reinforcement learning where an agent is asked to make sequential decisions, learning from feedback of an unknown environment \citep{strehl2007online}.

Furthermore, in time series forecasting, it is increasingly common to have access to \emph{multiple} signals with a \emph{hierarchical} structure at each time step. More precisely, we have access to a tree of time series, where the time series corresponding to a parent node is the sum of the time series of the children. This is the case, for example, in energy consumption, where the nodes would correspond to households, their parents to districts, and the top level to regions. Similarly, in a digital marketplace \citep[see, e.g.,][]{franses2011combining}, we are often interested in predicting demand for some products that can be grouped into different levels of categories (imagine we sell plum tomatoes, they belong to the category ``tomatoes'', which is a subcategory of ``vegetables''...). The idea is then to use this structure to improve the prediction at the leaf level. In fact, the aggregated signal at a parent node may be more stable and less noisy than the child time series. Exploiting this information can help improve the prediction of the child node compared to fitting a univariate model to each node.

These hierarchical forecasting problems can be recast as \emph{multivariate} online linear regression problems, which is the main topic of this paper (see Section~\ref{problem_statement_sec} for more details and a literature review on this problem).
Online linear regression is well studied in the univariate case, for which a reference algorithm is the Vovk-Azoury-Warmuth (VAW) algorithm, with its optimal $O(d \log(T))$ regret rate.
Instead, the literature on the multivariate case is more scarce.
To solve this problem, one can apply standard Online Convex Optimization (OCO) algorithms such as Online Gradient Descent (OGD), Follow The Regularized Leader (FTRL) or Online Newton Step (ONS).

However, a striking difference between these approaches and VAW is that they have no whatsoever theoretical guarantee for a global logarithmic regret bound. Note that this remains true even in the univariate case, as remarked for example in Section 7.11 of \citet{modern_ol_orabona_2022}. The main issue is that OGD and FTRL require strong convexity of the losses to go beyond sublinear rates (which we lack here), while ONS requires exp-concavity which holds under undesirable boundedness assumptions in this linear setup.

In short, online linear regression problems are much simpler than the worse OCO problem, so we should use an  algorithm that appropriately exploits this structure. 

We address this by introducing in Section \ref{S:multivaw} the algorithm MultiVAW, which can be seen as an extension of VAW 
to the multivariate case and which also enjoys a logarithmic regret rate.
In Section \ref{S:hierarchical forecasting application}, we apply MultiVAW to hierarchical forecasting problems, show that our method can solve these problems efficiently, and present numerical results in Section \ref{S:numerical experiments}.
In particular, we compare MultiVAW to an algorithm designed specifically for this forecasting problem, and show that it can be seen as a special case of MultiVAW, providing a new theoretical framework with relaxed assumptions for this competitor. All proofs are postponed to the appendix.

\paragraph{Notations.}
In this paper, we denote by $\N=\{1,2,\dots\}$ the set of positive integers, $\R=(-\infty,+\infty)$ the set of real numbers and $[T]=\{1,\dots,T\}$ for any $T\in\N$. We use uppercase letters to denote matrices, e.g., we denote the identity matrix by $I_d\in\R^{d\times d}$. Given a diagonalizable matrix $M\in\R^{d\times d}$ we denote its eigenvalues ordered in decreasing order as $\lambda_1(M)\geq\dots\geq\lambda_d(M)$. If a real matrix $M$ is positive semi-definite (resp. positive definite), we write $M\succeq 0$ (resp.  $M\succ 0$). We write $\norm{x}_2$ for the Euclidean norm of a vector $x \in \mathbb{R}^d$, and $\norm{x}_{\Lambda}=\sqrt{x^\top\Lambda x}$ for the norm induced by a matrix $\Lambda\succ 0$. For matrices, we write $\norm{\cdot}_F$ for the Frobenius norm. Finally, we define $\otimes$ as the Kronecker product operator between matrices, and note $\vecto(M) \in \mathbb{R}^{mn}$ the vectorization of a matrix $M \in \mathbb{R}^{m \times n}$.

\section{Problem Statement}
\label{problem_statement_sec}

\subsection{Multivariate Online Linear Regression}

We introduce the following sequential decision problem that we call \textit{multivariate online linear regression}, which involves the estimation of a vector of parameters in order to predict a vector of responses using a matrix of features.

For each time step $t\in[T]$:
\begin{enumerate}
    \item A matrix of features $X_t\in\R^{n_t\times d}$ is revealed.
    \item The decision-maker chooses $\theta_t\in\R^d$ and predicts $\hat{y}_t = X_t\theta_t$.
    \item A vector of responses $y_t\in\R^{n_t}$ is revealed and the decision-maker suffers the loss: $\norm{X_t\theta_t-y_t}_2^2$.
\end{enumerate}

The goal of the decision-maker is to minimize the total loss: $\sum_{t=1}^T \norm{X_t\theta_t-y_t}_2^2$. The performance is measured by the notion of \textit{regret} which is defined as follows:
\begin{equation}
    \label{regret_eq}
    R_T(\theta) = \sum_{t=1}^T \norm{X_t\theta_t-y_t}_2^2 - \sum_{t=1}^T \norm{X_t\theta-y_t}_2^2,
\end{equation}
where $\theta\in\R^d$ is a fixed vector called the \emph{competitor}.

\begin{example}[Univariate online linear regression]
\label{univariate_olr_ex}
Our framework is a simple multivariate extension of the univariate one \citep{olr_foster_1991}, which we recover by setting $n_t=1$ and $X_t=x_t^\top$ where $x_1,\dots,x_T\in\R^d$ denote the vectors of features.
\end{example}

\begin{remark}[Equivalent problems]
    \label{matrix_estimation_rmk}
    The main assumptions in our model are that predictions $\hat{y}_t$ are linear transformations of the features $X_t$ encoded by the parameters $\theta_t$, and that losses are defined by the squared Euclidean norm between predictions and responses.
    More generally, one could consider a model where the predictions have the form $\hat{Y}_t = V_t \Theta_t W_t$ where $V_t$ and $W_t$ are revealed features and $\Theta_t$ is a chosen parameter.
    Such structure naturally appears in online hierarchical forecasting (see Section \ref{application_ohf_sec}), and we verify in Appendix \ref{equivalent_model_appendix} that it fits in our framework.
\end{remark}

\begin{remark}[Links with OCO]
While our problem is clearly an online learning problem, we emphasize that it does not fit into the classical Online Convex Optimization framework. 
This is due to the fact that the losses $\Vert X_t \theta_t - y_t \Vert^2$ depend on the pair of data $(X_t,y_t)$ that are not simultaneously revealed.
\end{remark}
Below some examples of related problems in the literature.

\begin{example}[Online multiple-output regression]\label{Ex:framework online multiple output regression}
Consider the \textit{online multiple-output regression} framework of \citet{multi_olr_li_2018}, where at each time step $t\in[T]$, a vector of features $x_t\in\R^m$ is revealed, then, the decision-maker predicts $\hat{y}_t=\Theta_t x_t$ where $\Theta_t\in\R^{d\times m}$, after what the true response $y_t\in\R^d$ is revealed and the decision-maker updates its parameters (i.e. they define $\Theta_{t+1}$). Notice that in this framework the current feature $x_t$ is not incorporated in the current parameters $\Theta_t$. Instead, the decision-maker waits to see the pair $(x_t,y_t)$ before updating these. 
Apart from this difference, this model can be reduced to our framework. See Remark \ref{matrix_estimation_rmk} and Appendix \ref{equivalent_model_appendix}.
\end{example}

\begin{example}[Online state estimation]
The \textit{online state estimation} problem of \citet{online_state_estimation_cavraro_2021} corresponds to our framework where the pair $(X_t,y_t)$ is revealed at the beginning of time step $t$, the response writes $y_t=X_t\theta^*_t+\epsilon_t$ where $\epsilon_t\in\R^{n_t}$ is a noise vector and the goal is to track the true parameter $\theta^*_t\in\R^d$.
\end{example}

\subsection{Existing Algorithms and Results}

While linear regression is a standard topic in statistics, its online variant with deterministic sequences of features and responses was studied much more recently  and is mainly restricted to the univariate case (see Example~\ref{univariate_olr_ex}).

Univariate online linear regression has been first studied by \citet{olr_foster_1991} who gave a $O(d\log(dT))$ regret bound in the case where $y_t\in\{0,1\}$, $x_t\in[0,1]^d$ and parameters belong to the simplex. Then, \citet{vaw_vovk_2001} and \citet{vaw_azoury_2001} introduced the Vovk-Azoury-Warmuth (VAW) algorithm (also known as forward algorithm, or non-linear ridge regression algorithm) which is defined as
\begin{equation}
\tag{VAW}
\label{vaw_eq}
    \theta_t = \argmin_{\theta\in\R^d}  \sum_{s=1}^{t-1} (x_s^{\top}\theta-y_s)^2 + (x_t^{\top}\theta)^2 + \lambda \norm{\theta}_2^2,
\end{equation}
for some regularization parameter $\lambda>0$.
The first term $\sum_{s=1}^{t-1} (x_s^{\top}\theta-y_s)^2$ cumulates the past losses, the second term $(x_t^{\top}\theta)^2$ can be seen as a predictive loss that exploits the newly observed feature $x_t$, finally the last term $\lambda \norm{\theta}_2^2$ acts as a regularizer. 

Assuming that the features and responses are bounded, i.e., that there exist some constants $\bar{x}, \bar{y} > 0$ such that $\norm{x_t}_2 \leq \bar{x}$ and $|y_t|\leq \bar{y}$ for all $t\in [T]$, the \eqref{vaw_eq} algorithm achieves for all competitors $\theta\in\R^d$ a regret $R_T(\theta)$ that is bounded as follows (see \citet[Theorem 2]{olr_gaillard_2019} or \citet[Theorem 11.8]{prediction_cesa_2006}):
\begin{equation}
    \label{vaw_bound_eq}
    R_T(\theta) \leq \lambda \norm{\theta}_2^2 + d\bar{y}^2\log\left(1+\frac{T\bar{x}^2}{d\lambda}\right).
\end{equation}
To give some perspective on this bound, we precise that the minimax-optimal regret is of order $O(d\bar{y}^2\log(T))$, see Theorem 4 of \citet{olr_gaillard_2019}.

As far as we know, multivariate online linear regression attracted much less attention. 
The focus is usually made on the framework discussed in Example \ref{Ex:framework online multiple output regression}, for which dedicated algorithms have been proposed without any regret bounds, such as  iS-PLS  \citep{sparse_ls_mcwilliams_2010} or MORES \citep{multi_olr_li_2018}.

However, the setting of Example \ref{Ex:framework online multiple output regression} does not capture problems where the different components of the prediction $\hat{y}_t$ must satisfy linear relationships, as in the case of hierarchical forecasting. In fact, in the setting of Example~\ref{Ex:framework online multiple output regression}, each component of the response is treated independently (via the multiplication $\Theta_t x_t$) and one could therefore simply run parallel univariate algorithms. Such an approach is not allowed in our framework since predictions would then not satisfy the structural constraint $\hat{y}_t \in \image X_t$. In online hierarchical forecasting, this would produce forecasts that do not satisfy the summation constraints.

For all these reasons, we advocate for the use of a more general model for multivariate online linear regression and the design of methods tailored for this setting with strong theoretical guarantees.

\section{The MultiVAW Algorithm}\label{S:multivaw}
In order to solve our multivariate online linear regression problem, let us introduce a new algorithm which we call the Multivariate Vovk-Azoury-Warmuth (MultiVAW) algorithm. 
The latter allows to handle multidimensional responses but also more general forms of regularization compared to \eqref{vaw_eq}.

\subsection{Definition of MultiVAW}

Given a sequence of positive definite symmetric matrix $(\Lambda_t)_{t\in[T]}\subset\R^{d\times d}$ that we call \textit{regularization} matrices, the iterates of \eqref{multivaw_eq} are defined for $t\in[T]$ as follows:
\begin{equation}
\tag{MultiVAW}
\label{multivaw_eq}
    \theta_t = \argmin_{\theta\in\R^d}  \sum_{s=1}^{t-1} \norm{X_s\theta-y_s}_2^2 + \norm{X_t\theta}_2^2 + \norm{\theta}_{\Lambda_t}^2.
\end{equation}

Many algorithms can be seen as special instances of \eqref{multivaw_eq}. 
This is of course the case of \eqref{vaw_eq} which is recovered when considering the univariate problem (see Example \ref{univariate_olr_ex}) and taking the standard time-invariant regularization $\Lambda_t=\lambda I_d$ where $\lambda>0$.
Still in the univariate setting, when the features $(x_t)_{t\in[T]}$ are known beforehand (a model introduced by  \citet{minimax_olr_bartlett_2015}), \citet{olr_gaillard_2019} proposed an algorithm exploiting those features which is equivalent to \eqref{multivaw_eq} after taking the time-invariant regularization $\Lambda_t=\lambda\sum_{s=1}^T x_s x_s^\top$ for some $\lambda>0$. 
As far as we know, in the online linear regression literature time-varying regularizers have not been considered. Our framework allows to consider such, similarly to what is done with the FTRL method.

\subsection{Implementation and Complexity}
\label{impl_complexity_subsec}

The iterates of \eqref{multivaw_eq} are minimizers of a positive definite quadratic function, and as such can be computed efficiently. Indeed, by defining:
\begin{equation}
    \label{multivaw_close_eq_2}
    A_t = \Lambda_t + \sum_{s=1}^t X^\top_s X_s \text{ and } b_t=\sum_{s=1}^t X_s^\top y_s.
\end{equation}
we have that $\theta_t$ is uniquely determined by the linear equation $A_t\theta_t=b_{t-1}$ where $A_t\succ 0$, and can therefore be written:
\begin{equation}
    \label{multivaw_close_eq}
    \theta_t=A_t^{-1}b_{t-1}.
\end{equation}

Implementing \eqref{multivaw_eq} by computing sequentially $A_t$ and $b_t$ according to \eqref{multivaw_close_eq_2} and then, solving the corresponding system or inverting $A_t$, leads to a cost per time step that is of order $O(d^3 + n_t d^2)$. In Appendix \ref{special_reg_subsec} we show that in many cases this complexity can be improved.

\subsection{Regret Bounds for MultiVAW}
In this section, we state regret guarantees for \eqref{multivaw_eq}.
The following theorem is our main result.

\begin{theorem}
\label{main_theorem}
    Let $T\in\N$ and for every $t\in[T]$ let $X_t\in\R^{n_t\times d}$ and $y_t\in\R^{n_t}$. Define $\bar{y}=\sup_{t\in[T]}\norm{y_t}_2$ and consider the \eqref{multivaw_eq} algorithm ran with regularization matrices satisfying:
    \begin{equation*}
        0\prec \Lambda_1 \preceq \Lambda_2 \preceq \cdots  \preceq \Lambda_T.
    \end{equation*}
    For all $\theta\in\R^d$, we have:
    \begin{equation*}
        R_T(\theta) \leq \norm{\theta}_{\Lambda_T}^2
        +\bar{y}^2\sum_{i=1}^d \log\left(\frac{\lambda_i(A_T)}{\lambda_i(\Lambda_1)}\right),
    \end{equation*}
    where $A_T$ is defined by \eqref{multivaw_close_eq_2}.
\end{theorem}

In terms of proof techniques, these results do not follow from the analysis of FTRL but rather exploit directly the linear structure of the problem. We follow the analysis of \eqref{vaw_eq} (see e.g. \citet{olr_gaillard_2019}), and  incorporate techniques from the analysis of ONS. See Appendix \ref{S:proof of main result} for more details.

In the case of the standard time-invariant regularization $\Lambda_t=\lambda I_d$, we can derive from Theorem \ref{main_theorem} a simpler bound.

\begin{corollary} 
\label{standard_reg_cor}
Let assumptions of Theorem \ref{main_theorem} hold and set $\Lambda_t = \lambda I_d$ for some $\lambda > 0$, and $\bar{X} = \sup_{t\in[T]}\norm{X_t}_F$. 
For all $\theta\in\R^d$, we have:
\begin{equation*}
    R_T(\theta) \leq \lambda \norm{\theta}_2^2 + d\bar{y}^2 \log\left(1+\frac{T\bar{X}^2}{d\lambda}\right).
\end{equation*}
\end{corollary}

Theorem \ref{main_theorem} and Corollary \ref{standard_reg_cor} prove that \eqref{multivaw_eq} enjoys a logarithmic regret. Since we allow multivariate responses and more general time-varying regularization we recover as particular cases the usual \eqref{vaw_eq} regret bound \eqref{vaw_bound_eq} and other statements such as Theorem 2 of \citet{olr_gaillard_2019} or Theorem 11.8 of \citet{prediction_cesa_2006}.

In some sense, these results bridge the gap between the FTRL analysis (see e.g. Chapter 7 of \citet{modern_ol_orabona_2022}) which leads to $O(\sqrt{T})$ regret under milder assumptions and the \eqref{vaw_eq} analysis which leads to $O(\log{T})$ regret under stronger assumptions. Let us also mention that non-decreasing regularization is an assumption that commonly appears in the analysis of FTRL.

Notice that the bounds obtained do not depend explicitly on the dimension of the responses $n_t$. However, such a dependency may appear throughout the constants $\bar{y}$ and $\bar{X}$.

\section{Application: Online Hierarchical Forecasting}
\label{application_ohf_sec}
\label{S:hierarchical forecasting application}

\subsection{Introduction to Hierarchical Forecasting}
In hierarchical forecasting the goal is to forecast a \emph{hierarchical} time series, which is a multivariate time series $(y_t)_{t\in[T]}\subset \R^n$ that satisfies linear aggregation constraints \citep{amazon_rangapuram_2021}.
These constraints may model various structures such as a simple tree hierarchy, or more evolved grouped structures which involve crossed hierarchies \citep{fpp2_hyndman_2018}. A time series that satisfies those constraints is said to be \textit{coherent}. 
To formalize further coherency let us introduce the notion of \textit{summing matrix} through an example.

\begin{example}
    \label{hierarchy_ex}
    Consider an $8-$dimensional hierarchical time series $(y_t)_{t\in[T]}\subset \R^8$ subject to the constraints represented in Figure \ref{hierarchy_fig}.
    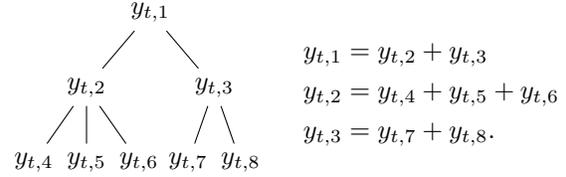
\begin{figure}[h]
    \centering
    \begin{minipage}[c]{0.2\textwidth}
    \begin{tikzpicture}
    [level distance=10mm,level 1/.style={sibling distance=17mm}, level 2/.style={sibling distance=7mm}]
      \node {$y_{t,1}$}
        child {node {$y_{t,2}$}
          child {node {$y_{t,4}$}}
          child {node {$y_{t,5}$}}
          child {node {$y_{t,6}$}}
        }
        child {node {$y_{t,3}$}
          child {node {$y_{t,7}$}}
          child {node {$y_{t,8}$}}
        };
    \end{tikzpicture}
    \end{minipage}
    \begin{minipage}[c]{0.25\textwidth}
    \vspace{-10pt}
     \begin{align*}
        y_{t,1} &= y_{t,2}+y_{t,3}\\
        y_{t,2} &= y_{t,4}+y_{t,5}+y_{t,6}\\
        y_{t,3} &= y_{t,7}+y_{t,8}.
    \end{align*}
    \end{minipage}
    \caption{An example of hierarchical time series.}
    \label{hierarchy_fig}
    \end{figure}
    
    In this example, the components of $(y_t)_{t\in[T]}$ can be split into aggregated series: $(y_{t,i})_{t\in[T]}$ for $i=1,2,3$, represented as internal nodes, and bottom-level series: $(y_{t,i})_{t\in[T]}$ for $i=4,\dots,8$, represented as leaf nodes. Aggregated series are therefore sums of bottom-level series, thus, the whole series $(y_t)_{t\in[T]}$ can be represented as a linear transformation of bottom-level series. Formally, $y_t=S b_t$ where $b_t=(y_{t,i})_{i=4,\dots,8}$ and $S$ is defined below:
    \begin{equation*}
        S=\begin{pmatrix}
        \begin{matrix} 1 & 1 & 1 & 1 & 1\\ 1 & 1 & 1 & 0 & 0 \\ 0 & 0 & 0 & 1 & 1 \end{matrix} \\ I_5
        \end{pmatrix}
    \end{equation*}
    Such a matrix $S$ encode the summation constraints and is called the \textit{summing matrix}. Coherent time series are characterized by $y_t\in\image S$ for all $t\in[T]$.
\end{example}

To solve the hierarchical forecasting problem, a two-step approach is typically followed \citep[see, e.g.,][]{temporal_hierarchies_athanasopoulos_2017,taieb2017coherent}. 
In a first step, \textit{base forecasts} are produced using some features, but without trying to be coherent. 
Only in a second step, these base forecasts are transformed into coherent ones by a so-called \textit{reconciliation method}.
For an introduction to this approach, we refer the reader to \citet{fpp2_hyndman_2018}. We also mention that some deep learning architectures that ensure hierarchical coherence have been proposed \citep{mancuso2021machine,sagheer2021deep}.
In our paper, we do not follow this two-step approach and instead take a direct approach using an online framework. To our knowledge, the online version has received little to no attention, with the exception of the work of \citet{ohf_bregere_2022}. 

\subsection{Online Hierarchical Forecasting}

In online hierarchical forecasting the features and responses are revealed sequentially. 
Following \citet{ohf_bregere_2022}, we assume that there is a vector of features $x_t\in\R^m$ available at the beginning of time step $t\in[T]$.
However, in this work, we allow the linear constraints to be arbitrary and time-varying. 
Thus, we consider an arbitrary sequence of summing matrices $S_t\in \R^{n_t\times d}$ which are also made available sequentially at the beginning of time step $t\in[T]$. 
Coherent time series $(\hat y_t)_{t\in[T]}$ are then characterized by $\hat y_t\in \image S_t$ for all $t\in[T]$. 
Our goal is to learn a linear relationship between the features $x_t$ and the responses $y_t$ which are arbitrary vectors of $\R^{n_t}$ known at the \textit{end} of time step $t$. 
Any linear function from the feature space $\R^m$ to the space of coherent time series $\image S_t$ have the form $x \mapsto S_t \Theta_t x$ where $\Theta_t\in\R^{d\times m}$ is a matrix of parameters. 
Thus, considering the squared Euclidean norm metric, our problem fit in our original framework (see Lemma \ref{ohf_to_molr_lemma} in Appendix \ref{equivalent_model_appendix} for more details).

To sum up, we consider the following sequential decision problem we call \textit{online hierarchical forecasting} (OHF). For each time step $t\in[T]$:
\begin{enumerate}
    \item A summing matrix $S_t\in\R^{n_t \times d}$ and a feature vector $x_t\in\R^m$ are revealed.
    \item The decision-maker predicts $\hat{y}_t\in\image(S_t)$.
    \item The response $y_t\in \R^{n_t}$ is revealed and the decision-maker suffers the loss: $\norm{\hat{y}_t - y_t}_2^2$.
\end{enumerate}

\subsection{Related Work and MetaVAW}

We now discuss the model and results of \citet{ohf_bregere_2022} who consider a similar problem\footnote{In the following we will mainly refer to the preprint version \citep{ohf_arxiv_bregere_2020} because it has extended results.}.
Before diving into their results, let us mention that they model the summation constraints in a different but equivalent way: coherent signals lie in the kernel of a matrix $K$ rather than in the image of a summing matrix $S$. Their representation is usually more compact but less common in the literature of hierarchical forecasting.

Assuming that the constraints are time-invariant, i.e., $n_t=n$ and $S_t=S$ for all $t\in[T]$, they designed an online meta-algorithm (See Meta-Algorithm 1 in \citet{ohf_arxiv_bregere_2020}) that follows the common two-step approach of hierarchical forecasting. 
Base forecasts are produced by $n$ independent instances of a \emph{base online algorithm} which is run to predict the univariate series $(y_{t,1})_{t\in[T]}, \dots, (y_{t,n})_{t\in[T]}$ using the whole vector of features $x_t$ each time, after which a reconciliation is performed through a projection onto the set of coherent signals (also known as OLS reconciliation).

Several cases of base online algorithms are discussed in \citet{ohf_arxiv_bregere_2020} such as: \eqref{vaw_eq}, BOA \citep{boa_wintenberger_2017} or ML-Poly \citep{phd_gaillard_2015}. 
In the sequel we will only focus on \eqref{vaw_eq}, and we will call the resulting meta-algorithm \emph{MetaVAW}. 
Let us provide a closed-form expression for this algorithm.

\begin{proposition}
    \label{metavaw_close_prop}
    Consider an OHF problem with time-invariant constraints defined by a  summing matrix $S\in\R^{n\times d}$. 
    For a regularization parameter $\lambda>0$, the predictions of MetaVAW can be written as:
    \begin{equation}
        \label{metavaw_close_eq}
        \hat{y}_t = P_S\left(\sum_{s=1}^{t-1} y_s x_s^\top\right)\left(\lambda I_m + \sum_{s=1}^t x_s x_s^\top\right)^{-1}x_t,
    \end{equation}
    for all $t\in[T]$, where $P_S=SS^\dagger$ is the projection matrix into $\image S$.
\end{proposition}

Regarding the guarantees for MetaVAW, \citet{ohf_arxiv_bregere_2020} provide a regret bound that scales like $O(n^2\log(T))$ under specific assumptions which we discuss below in Remark \ref{on_ohf_ass_rmk}.

\subsection{MultiVAW for Online Hierarchical Forecasting}
As detailed in Lemma \ref{ohf_to_molr_lemma} of Appendix \ref{equivalent_model_appendix} we can reduce the OHF problem to multivariate online linear regression. Thus, we can resort to \eqref{multivaw_eq} to solve the OHF problem.
Our method specialized to this setting is dubbed MultiVAW-OHF and it is summarized in Algorithm \ref{multivaw_for_ohf}.

\begin{algorithm}[h]
\caption{MultiVAW-OHF}\label{multivaw_for_ohf}
\begin{algorithmic}
\STATE \textbf{Parameters:} $dm\times dm$ regularization matrices
\STATE $0\prec \Lambda_0=\Lambda_1 \preceq \Lambda_2 \preceq \dots \preceq \Lambda_T$
\STATE \textbf{Initialization} Let $A_0 = \Lambda_0 $, $b_0=0$
\FOR{$t\in[T]$}
\STATE Observe $S_t\in\R^{n_t\times d}$ and $x_t \in \R^m$
\STATE $X_t = x_t^\top \otimes S_t$
\STATE $A_t = A_{t-1}-\Lambda_{t-1}+\Lambda_t+X_t^\top X_t$
\STATE $\theta_t=A_t^{-1}b_{t-1}$
\STATE Predict $\hat y_t = X_t \theta_t$
\STATE Observe $y_{t} \in \R^{n_t}$
\STATE Update $b_t = b_{t-1}+X_t^\top y_t$
\ENDFOR
\end{algorithmic}
\end{algorithm}

Interestingly, we show that our competitor MetaVAW corresponds exactly to MultiVAW-OHF with a specific regularizer choice.

\begin{proposition}
    \label{meta_is_multi_prop}
    Consider an OHF problem with time-invariant constraints defined by an injective summing matrix $S\in\R^{n\times d}$, and let $\lambda>0$.
    Then MetaVAW with regularization $\lambda$ generates the exact same predictions than MultiVAW-OHF with $\Lambda_t=\lambda I_m\otimes (S^\top S)$ for all $t\in[T]$.
\end{proposition}

In Proposition \ref{meta_is_multi_prop} we assume  time-invariant constraints which is very standard in the literature of hierarchical forecasting and the injectivity of $S$ which is guaranteed whenever it has the form $S=(*, I_d)^\top$.

Under these mild conditions, Proposition \ref{meta_is_multi_prop} sheds a new light on MetaVAW. By observing that $\norm{\vecto(\Theta)}_{\lambda I_m\otimes (S^\top S)}^2=\lambda\norm{S\Theta}_F^2$ we can rewrite its predictions as $\hat{y}_t=S\Theta_t x_t$ where $\Theta_t$ is defined as:
\begin{equation*}
    \Theta_t=\argmin_{\Theta\in\R^{d\times m}}{\sum_{s=1}^{t-1}\norm{S\Theta x_s-y_s}_2^2+\norm{S\Theta x_t}_2^2+\lambda\norm{S\Theta}_F^2}
\end{equation*}

Compared to MultiVAW-OHF with standard regularization where $\Lambda_t=\lambda I_{dm}$, we see that the regularization term of MetaVAW is $\lambda\norm{S\Theta}_F^2$ instead of $\lambda\norm{\Theta}_F^2$.

\subsection{Regret Guarantees for MultiVAW-OHF}

It is straightforward to reformulate Theorem \ref{main_theorem} in the context of OHF to obtain regret bounds. In this setting, the regret \eqref{regret_eq} for a competitor $\Theta \in\R^{d\times m}$ has the form:
\begin{equation*}
    R_T(\Theta) = \sum_{t=1}^T \norm{\hat{y}_t-y_t}_2^2 - \sum_{t=1}^T \norm{S_t\Theta x_t-y_t}_2^2.
\end{equation*}

In particular, we can derive the following regret bounds for MultiVAW-OHF with standard regularization as well as for MetaVAW.

\begin{corollary}
    \label{ohf_bounds_cor}
    Consider an OHF problem and $\lambda>0$. Define $\bar{x} = \max_{t\in[T]}\norm{x_t}_2$ and $\bar{S}=\max_{t\in[T]}\norm{S_t}_F$.
    \begin{enumerate}
        \item MultiVAW-OHF with standard time-invariant regularization $\Lambda_t=\lambda I_{dm}$ has a regret $R_T(\Theta)$ bounded, for all $\Theta\in\R^{d\times m}$, by:
        \begin{equation}
            \label{multivaw_ohf_bound_eq}
            \lambda \norm{\Theta}_F^2 + dm \bar{y}^2\log\left(1+\frac{T\bar{x}^2\bar{S}^2}{dm\lambda}\right).
        \end{equation}
        \item Assuming time-invariant constraints defined by an injective summing matrix $S\in\R^{n\times d}$, MetaVAW has a regret $R_T(\Theta)$ bounded, for all $\Theta\in\R^{d\times m}$, by:
        \begin{equation}
            \label{metavaw_bound_eq}
            \lambda \norm{S\Theta}_F^2 + dm \bar{y}^2\log\bigg(\Big(1+\frac{T\bar{x}^2}{m\lambda}\Big)\frac{\norm{S}_F^2}{d\lambda_d(S^\top S)}\bigg).
        \end{equation}
    \end{enumerate}
\end{corollary}

\begin{remark}[On the assumptions]
\label{on_ohf_ass_rmk}
    Corollary \ref{ohf_bounds_cor} provides logarithmic regret bounds for online hierarchical forecasting under much weaker assumptions than in \citet{ohf_arxiv_bregere_2020}. First, we allow time-varying summing matrices. Second, we do not require the responses $(y_t)_{t\in[T]}$ to be coherent. This allows us to take into account errors in the observed signal which may not always satisfy the linear constraints. This kind of errors may happen in real-world applications such as energy load forecasting.
    Finally, \citet{ohf_arxiv_bregere_2020} assume further that the features and responses have the same dimension, i.e. $m=n$, an assumption we do not need here.
\end{remark}

\begin{remark}[Comparing MetaVAW and MultiVAW-OHF]
In terms of computational cost, we observe that MultiVAW-OHF is in general more expensive than MetaVAW. This is not surprising, given that MultiVAW-OHF is a more general algorithm that, unlike MetaVAW, is able to handle time-varying constraints. A more detailed comparison of the complexity of these two algorithms is proposed at the end of Appendix \ref{special_reg_subsec}.

Now, in terms of regret, we can compare in more practical settings the bounds obtained in Corollary \ref{ohf_bounds_cor}. It happens that in most of the use cases involving a single hierarchy as in Example \ref{hierarchy_ex} the bound \eqref{multivaw_ohf_bound_eq} is smaller than \eqref{metavaw_bound_eq}.
In other words, the guarantees obtained for MultiVAW-OHF with standard regularization are better than those obtained for MetaVAW. This is detailed in the following example.
\end{remark}

\begin{example}
    Assume that we are dealing with a single hierarchy as in Example \ref{hierarchy_ex} where at least two bottom-level nodes have the same ancestors. In such a case, the common way of defining the summing matrix is by setting $S=(\tilde{S}^\top, I_d)^\top$ where $\tilde{S}\in\R^{(n-d)\times d}$. This means that each signal $y_t\in\R^n$ is written $y_t=(a_t^\top, b_t^\top)^\top$, where $a_t\in\R^{n-d}$ are aggregated signals and $b_t\in\R^d$ are bottom-level signals. By assumption, there exists two columns of $\tilde{S}$ that are equal. This leads to $\lambda_d(\tilde{S}^\top \tilde{S})=0$ and therefore $\lambda_d(S^\top S)=1$ which implies that each one of the two terms of \eqref{multivaw_ohf_bound_eq} is smaller than the corresponding term of \eqref{metavaw_bound_eq}.
\end{example}

\section{Numerical Experiments}\label{S:numerical experiments}

In this section, we compare numerically different algorithms for online hierarchical forecasting using real-world datasets.

\subsection{Algorithms}
Four algorithms are compared: MultiVAW-OHF, MetaVAW, FTRL and OGD.
We use and standard time-invariant regularization for both MultiVAW-OHF ($\Lambda_t = \lambda I_{dm}$) and FTRL which predicts $\hat{y}_t=S \Theta_t x_t$ where:
\begin{equation*}
    \Theta_t = \argmin_{\Theta\in\R^{d\times m}}  \sum_{s=1}^{t-1} \norm{S\Theta x_s-y_s}_2^2 + \lambda\norm{\Theta}_{F}^2.
\end{equation*}
The difference between FTRL and MultiVAW-OHF is the predictive term $\norm{X_t\theta}_2^2$ which is discarded for the former. Regarding OGD, its iterates write:
\begin{equation*}
    \theta_{t+1} = \Proj_{[-M,M]^{dm}}\left(\theta_t - \eta 2X_t^\top(X_t\theta_t-y_t)\right)
\end{equation*}
where we set $\theta_1=0$, $M\geq 0$ a large enough constant and $\eta = 10^{-9}/\lambda$. The Euclidean projection operator is denoted by $\Proj$, whereas, $X_t$, $y_t$, $\theta_t$ has already been defined in the context of OHF.

\begin{figure}[!ht]
    \includegraphics[width=0.5\textwidth]{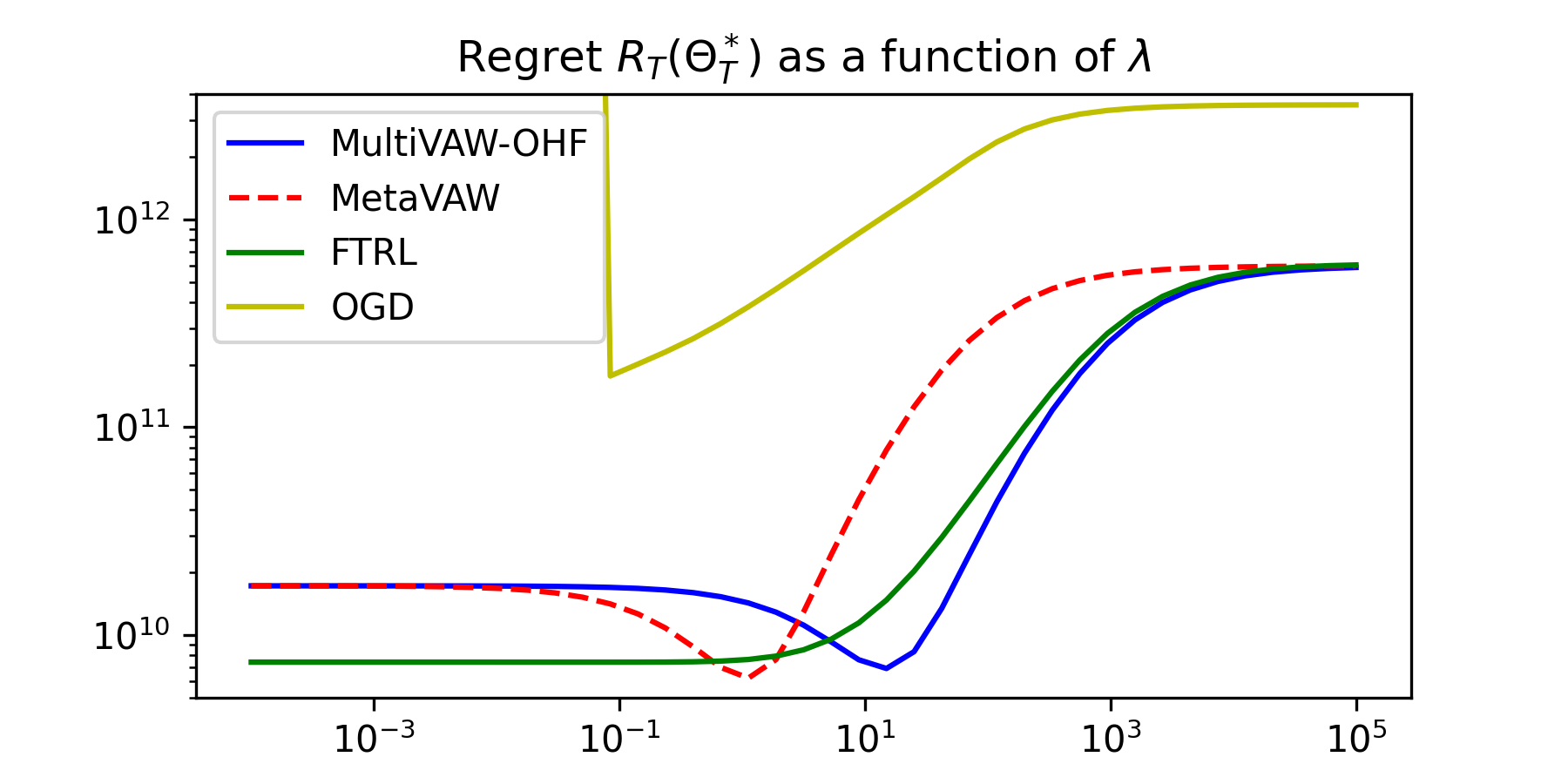}
    \includegraphics[width=0.5\textwidth]{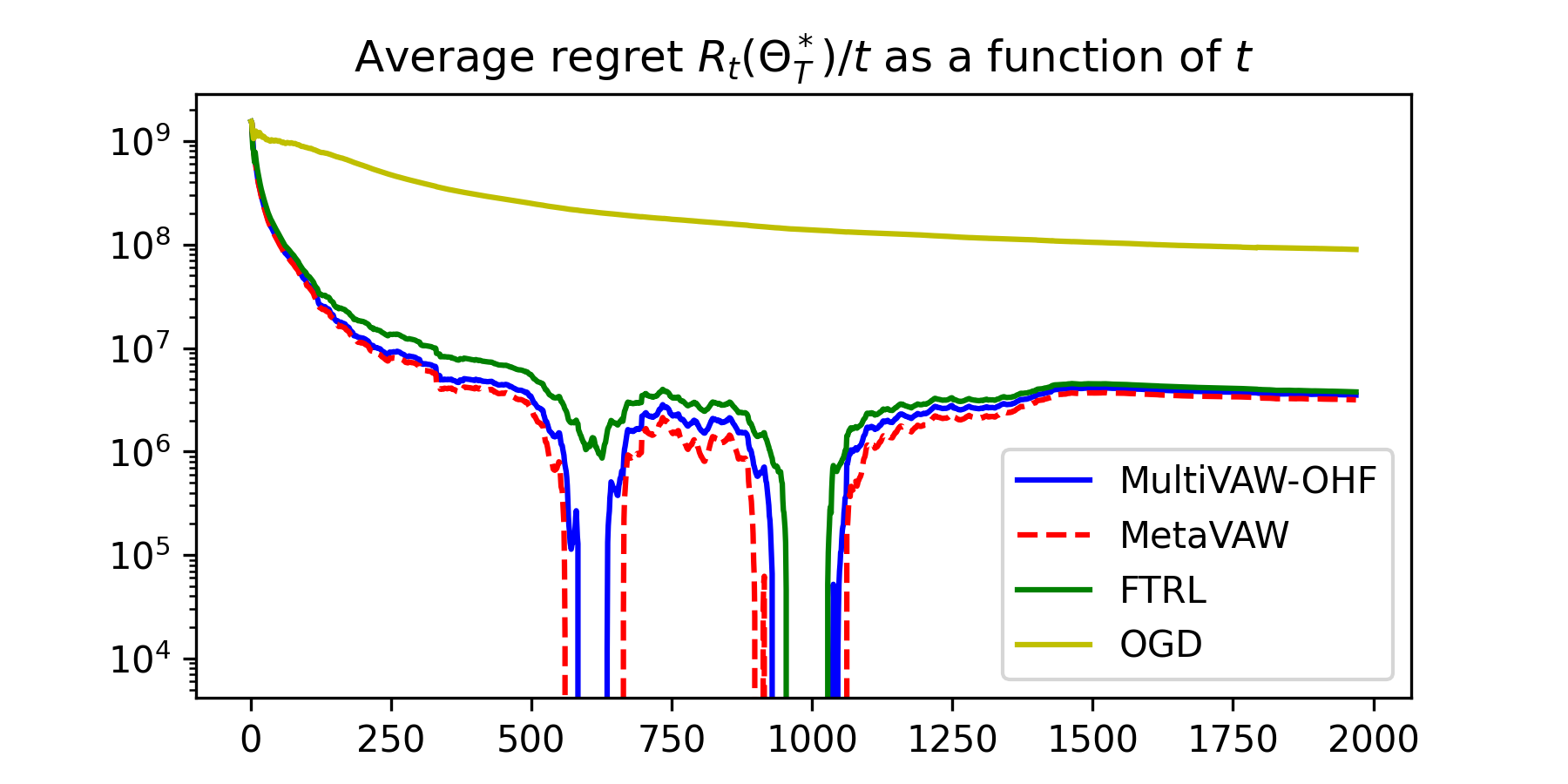}
    \includegraphics[width=0.5\textwidth]{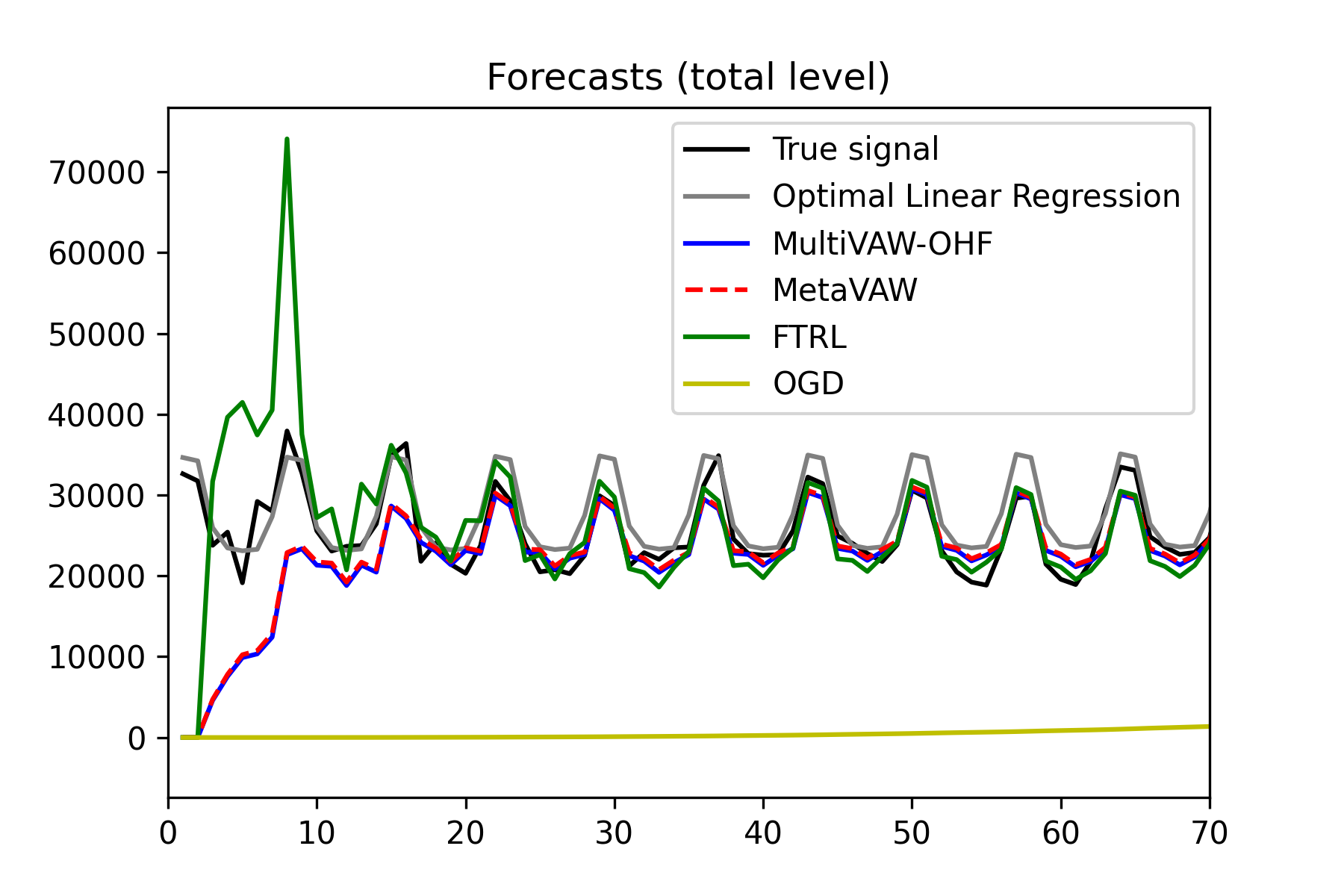}
    \includegraphics[width=0.5\textwidth]{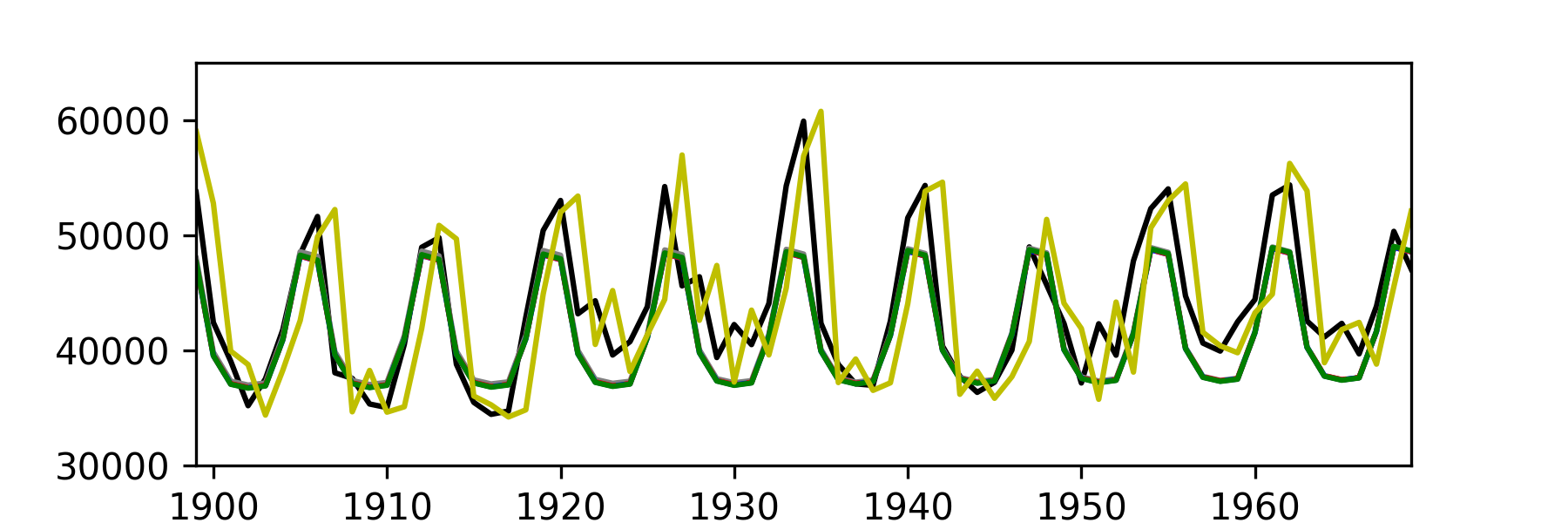}
    \caption{\texttt{M5} dataset}
    \label{tourismsmall_fig}
\end{figure}

\begin{figure}[!ht]
    \includegraphics[width=0.5\textwidth]{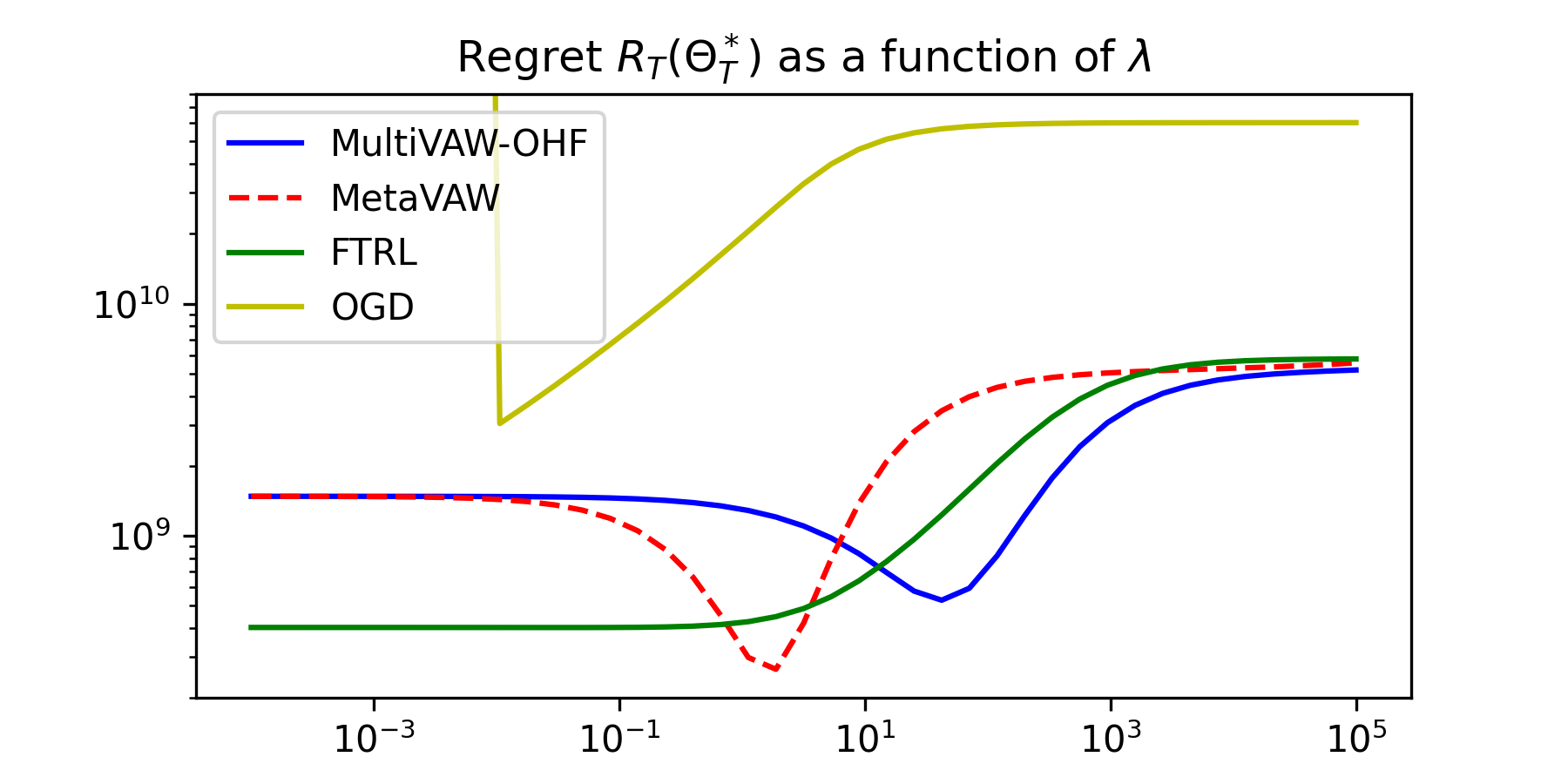}
    \includegraphics[width=0.5\textwidth]{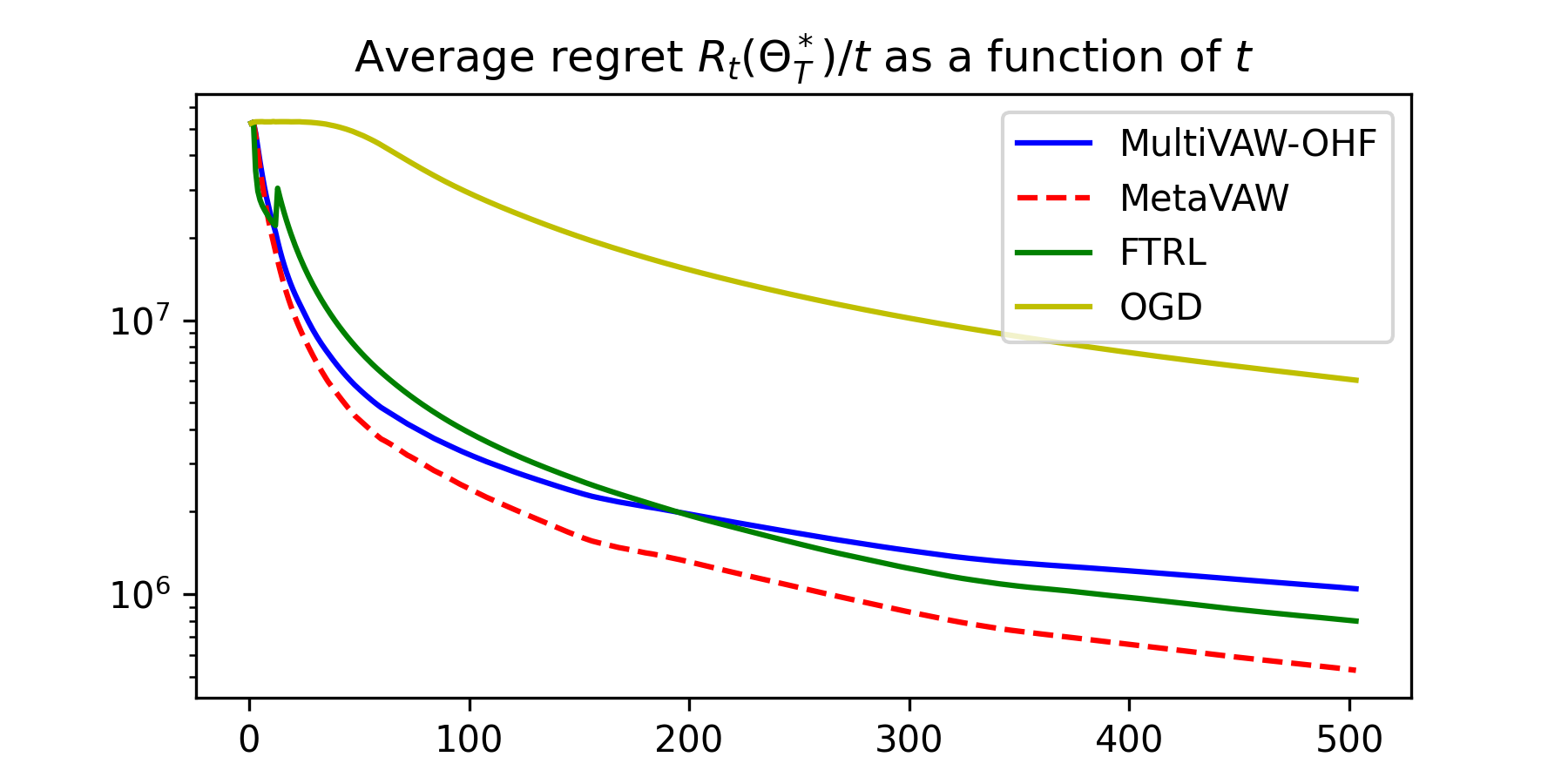}
    \includegraphics[width=0.5\textwidth]{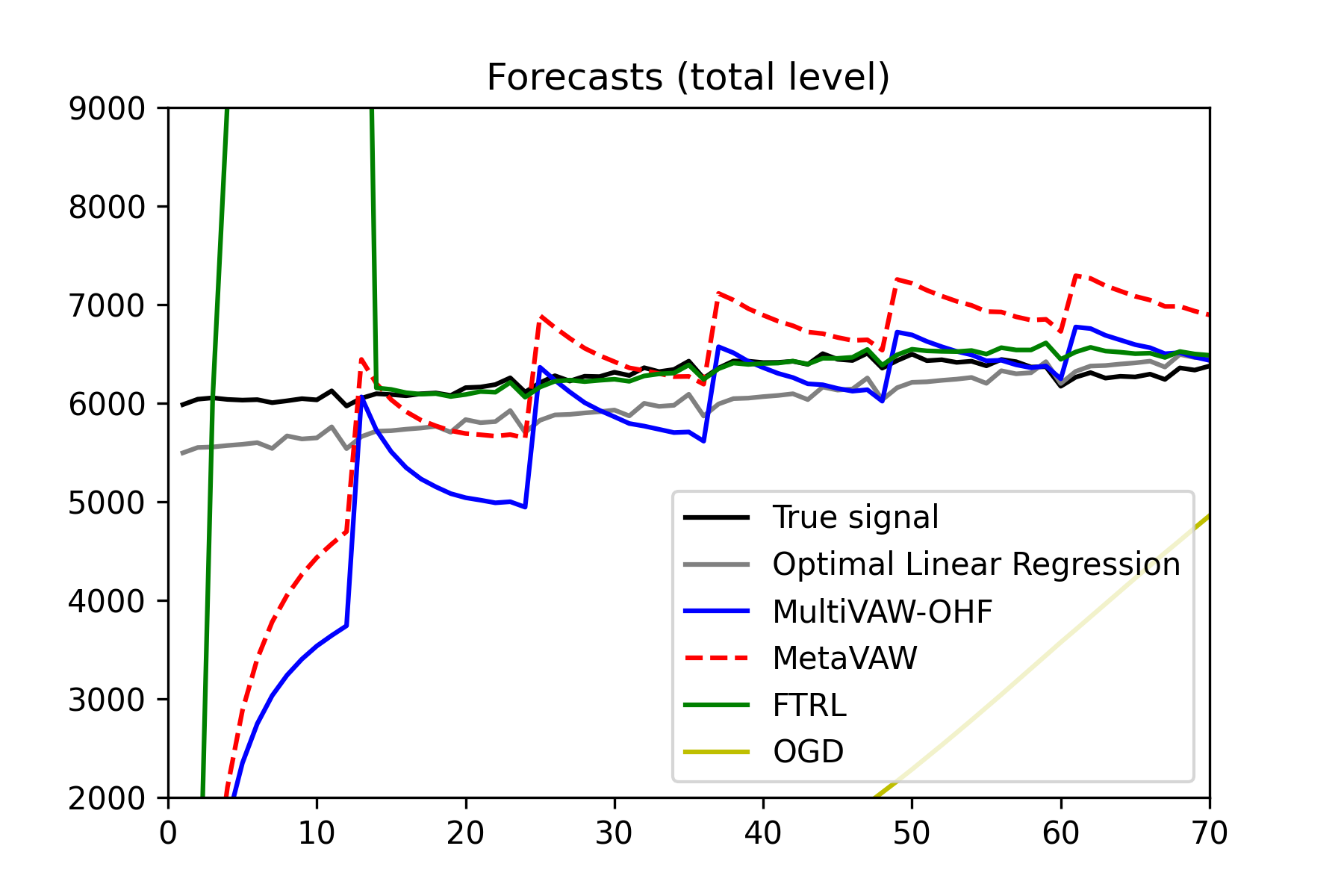}
    \includegraphics[width=0.5\textwidth]{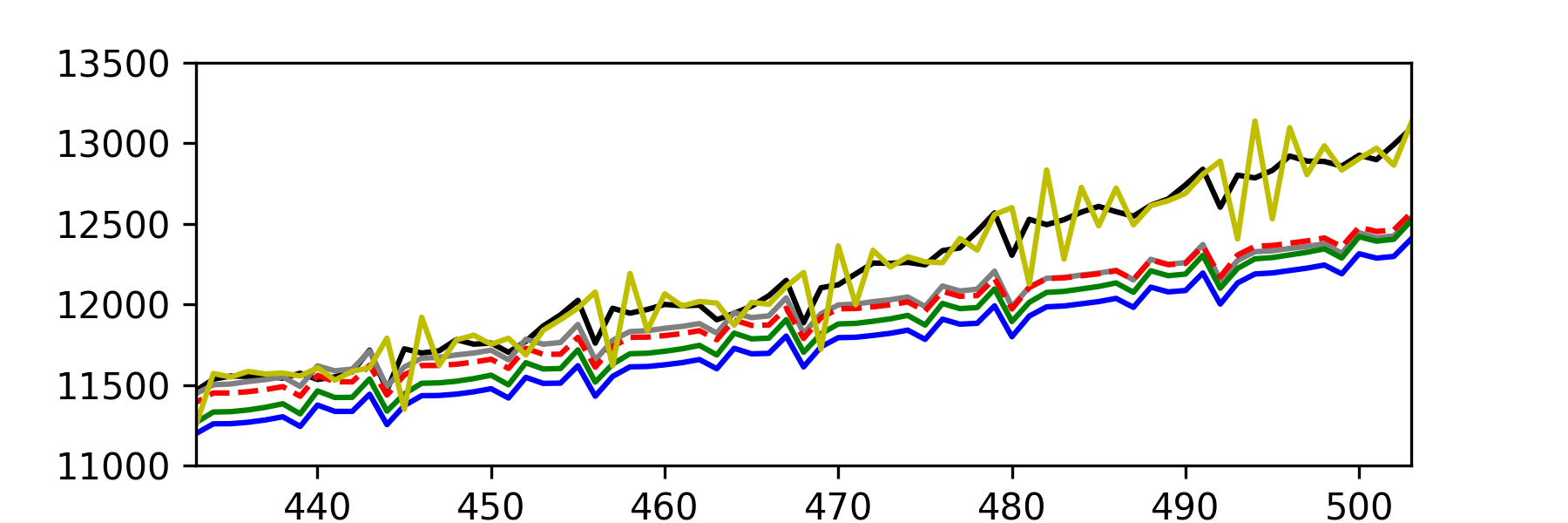}
    \caption{\texttt{Labour} dataset}
    \label{labour_fig}
\end{figure}

\subsection{Datasets and Feature Engineering}
Two datasets are considered: \texttt{M5} and \texttt{Labour}, which are available in the \href{https://github.com/Nixtla/datasetsforecast}{\texttt{datasetsforecast}} package under the MIT license. For \texttt{M5} we extracted a sub-hierarchy featuring $n=14$ nodes with $d=10$ bottom-level series, it is daily sampled with $T=1969$ and features a weakly seasonality and a slight upward trend. On the other hand, \texttt{Labour} consist of $n=57$ nodes and $d=32$ bottom-level series, it is monthly sampled with $T=503$ and features a yearly seasonality and a stronger upward trend.

Regarding the choice of the features, we included for both datasets the time step $t$ by setting $x_{t,1}=t$, then, to capture seasonality, we added for \texttt{M5} the one-hot encoding of the ``day of the week'' so that $m=8$, and for \texttt{Labour} the one-hot encoding of the ``month of the year'' so that $m=13$.

\subsection{Results}
Figures \ref{tourismsmall_fig} and \ref{labour_fig} show the performances of the algorithms. 
For each dataset, we first show the regret $R_T(\Theta^*_T)$ where $\Theta^*_T = \argmax_\Theta R_T(\Theta)$ as a function of the regularization parameter $\lambda$.
This allows to see the sensitivity of the algorithm with respect to $\lambda$, but also allows for a fair comparison between algorithms.
Then, we show the average regret $R_t(\Theta^*_T)/t$ as a function of $t$ for each algorithm (with a parameter $\lambda$ optimally chosen from the previous plot).
Finally, we show the associated forecasts on the first time steps, then, on the last time steps, at the total level in addition to the true responses and the forecasts associated to the optimal linear regression $\Theta^*_T$.

In both datasets, OGD has the worst performances and MetaVAW the best ones. This proves first that in this context, regularization-based methods such as FTRL or VAW-based methods should be preferred to OGD. This also suggests that MetaVAW, as the most specialized method, is able to achieve the best performances when tuned optimally.

Surprisingly, FTRL is performing very well for a generic method, outperforming other methods when $\lambda \sim 0$. Let us also mention that in this regime, the behavior of MultiVAW-OHF and MetaVAW becomes similar which is no surprise according to Proposition \ref{meta_is_multi_prop}.

All the methods seems to exhibit a decreasing average regret which is the standard theoretical requirement in online learning. Of course, the rates are different, OGD being much slower than its competitors.

In terms of quality of the forecasts, we see in the first time steps that OGD is very slow to fit the true signal whereas FTRL lacks robustness. The latter phenomenon was already observed in the univariate stochastic setting by \citet{sto_olr_ouhamma_2021}, and was part of their justification about why VAW-based methods should be preferred to FTRL. On the last time steps, we see that OGD essentially behaves like a shifted version of the true signal as it responds to the error it just observed. Its competitors on the other hand are able to track much more efficiently the optimal linear forecast associated to $\Theta_T^*$.

In terms of computation time, we observe that MetaVAW runs faster than the others. Of course this observation depends on the implementation, but at least we can say that we found quite easy to implement MetaVAW efficiently by exploiting the closed-form formula \eqref{metavaw_close_eq}.

All these experiments suggest that in specialized setups with time-invariant summing matrix, an optimally tuned MetaVAW algorithm allows for the best performances. However, when the summing matrices are time-varying, MetaVAW is no longer applicable, making MultiVAW-OHF a viable algorithm. 
Finally, our experiments showed that FTRL  is a rather strong and flexible baseline for OHF problems, besides a certain lack of robustness in the forecasts.

\section{Conclusion}

In this paper, we have addressed the problem of multivariate online linear regression. To this end, we proposed \eqref{multivaw_eq}, a generalization of the \eqref{vaw_eq} algorithm \citep{vaw_vovk_2001, vaw_azoury_2001}, which handles the multivariate setting and more general forms of regularization. Strong theoretical guarantees in the form of logarithmic regret bounds have been derived for \eqref{multivaw_eq}, generalizing in this respect bounds known for \eqref{vaw_eq} in the univariate case.

Our results have been applied to the online hierarchical forecasting problem (OHF), where we recovered the MetaVAW algorithm of \citet{ohf_bregere_2022} as an instance of \eqref{multivaw_eq} with a particular regularization. Our approach allowed us to relax some assumptions made for its analysis.

We also conducted numerical experiments in the context of OHF to compare instances of MultiVAW and other baselines. MetaVAW being the most specialized algorithms to our simulations it emerged naturally as the best performing method.

As future work, it would be interesting to study how the special structure of summing matrices impacts the regret bounds. Indeed, the results we presented for OHF such as Corollary \ref{ohf_bounds_cor} are agnostic to the particular structure that usual summing matrices exhibit, and it would be interesting to know how that structure can impact the rates.

\section*{Impact Statement}
This paper presents work whose goal is to advance the field of Machine Learning. There are many potential societal consequences of our work, none which we feel must be specifically highlighted here.


\bibliography{biblio}
\bibliographystyle{icml2024}

\newpage
\appendix
\onecolumn


\section{An Equivalent Model}
\label{equivalent_model_appendix}

To make easier the link with online hierarchical forecasting and show the generality of our framework we introduce in this appendix an \textit{alternative} model that is in fact equivalent to multivariate online linear regression. It involves the estimation of a matrix of parameters in order to predict a matrix of responses using two matrices of features. 

More precisely, for each time step $t\in[T]$:
\begin{enumerate}
    \item Two matrices of features $W_t\in\R^{m\times k_t}$ and $V_t\in\R^{n_t\times d}$ are revealed.
    \item The decision-maker chooses $\Theta_t\in\R^{d\times m}$ and predicts $\hat{Y}_t = V_t \Theta_t W_t$.
    \item A matrix of responses $Y_t\in\R^{n_t\times k_t}$ is revealed and the decision-maker suffers the loss $\norm{V_t\Theta_t W_t-Y_t}_F^2$.
\end{enumerate}

\begin{proposition} 
    \label{equivalent_models}
    This alternative model is equivalent to multivariate online linear regression.
\end{proposition}
\begin{proof}
    First, it is clear that multivariate online linear regression is a particular case of the alternative model: simply take $W_t=1$, $V_t=X_t$, $\Theta_t=\theta_t$ and $Y_t=y_t$. 
    
    Secondly, the converse holds, the alternative problem can be seen as an instance of multivariate online linear regression. We can formally see this connection by setting $X_t=(W_t^\top \otimes V_t)\in\R^{n_t k_t\times dm}$, $\theta_t=\vecto(\Theta_t)\in\R^{dm}$ and $y_t=\vecto(Y_t)\in\R^{n_t k_t}$ so that:
    \begin{align*}
    \norm{V_t\Theta_t W_t-Y_t}_F^2 &= \norm{\vecto(V_t\Theta_t W_t)-\vecto (Y_t)}_2^2 \\
    &= \norm{(W_t^\top \otimes V_t)\vecto(\Theta_t)-\vecto(Y_t)}_2^2 \\
    &= \norm{X_t\theta_t-y_t}_2^2.
    \end{align*}
\end{proof}

\begin{lemma}
    \label{ohf_to_molr_lemma}
    The OHF problem (see Section \ref{application_ohf_sec}) is a  multivariate online linear regression problem (see Section \ref{problem_statement_sec}) with $X_t = x_t^\top\otimes S_t \in\R^{n_t\times dm}$ and $\theta_t=\vecto(\Theta_t)\in \R^{dm}$.
\end{lemma}
\begin{proof}
    OHF corresponds to the alternative model with $k_t=1$, $W_t=x_t$ and $V_t=S_t$. Therefore, the reduction we detailed in the proof Proposition \ref{equivalent_models} applies and defining $X_t$ and $\theta_t$ as claimed reduces OHF to multivariate online linear regression.
\end{proof}

\section{Improved Complexity}
\label{S:computational cost matrix estimation}
\label{special_reg_subsec}

In this section we discuss improved complexity results for \eqref{multivaw_eq} in some special cases.

\subsection{Original Model}
In multivariate online linear regression (see Section \ref{problem_statement_sec}) consider \eqref{multivaw_eq} using a time-invariant regularization, i.e. $\Lambda_t=\Lambda_1$. In such a case, the recursive formula for $A_t$ boils down to $A_t = A_{t-1} + X_t^\top X_t$.
Thus, in a regime where $n_t$ is small compared to $d$, it is preferable not to invert $A_t$ directly, but rather to use the Woodburry formula to express $A_t^{-1}$ in terms of $A_{t-1}^{-1}$ and $X_t$ as follows:
\begin{equation*}
A_t^{-1} = A_{t-1}^{-1}-A_{t-1}^{-1}X_t^\top\left(I_{n_t}+X_tA_{t-1}^{-1}X_t^\top\right)^{-1}X_tA_{t-1}^{-1}.
\end{equation*}
This leads to a lower cost per iteration of $O(n_t d^2 + n_t^2 d + n_t^3)$ instead of $O(d^3+n_t d^2)$. In particular, in the univariate case, we recover the $O(d^2)$ cost of \eqref{vaw_eq} discussed by \citet[Section 11.8]{prediction_cesa_2006}.

\subsection{Alternative Model}

More generally, this computation cost can be very problem-dependant. As an example, consider the alternative model problem of Appendix \ref{equivalent_model_appendix}. By reducing this problem to our original framework we would naively obtain a complexity like $O(m^3d^3+n_t k_t m^2 d^2)$, or in the time-invariant regularization case $O(n_t k_t m^2 d^2 + n_t^2 k_t^2 m d + n_t^3 k_t^3 )$ using the Woodburry formula, so, in both cases, a polynomial of degree six.

However, by exploiting the structure of the problem and the regularization this cost can be can be drastically improved in some cases. The following proposition provides a setting where the iterates of \eqref{multivaw_eq} enjoy a simple closed-form expression that leads to an improved complexity.

\begin{proposition}
    \label{special_reg_prop}
    Consider the alternative model of Appendix \ref{equivalent_model_appendix} and assume further that the matrices $V_t$ are time-invariant, i.e., $n_t=n$ and $V_t=V$ for all $t\in[T]$, with $V$ injective.
    Let $\tilde{\Lambda}\in\R^{m\times m}$ be a fixed symmetric positive definite matrix.
    
    Then, the iterates of \eqref{multivaw_eq} ran with the regularization $\Lambda_t =\tilde{\Lambda}\otimes (V^\top V)$ have the following closed-form expression:
    \begin{equation}
        \label{special_reg_close_eq}
        \Theta_t = (V^\top V)^{-1} V^\top \Big(\sum_{s=1}^{t-1}Y_s W_s^\top\Big)\Big(\tilde{\Lambda}+\sum_{s=1}^t W_s W_s^\top\Big)^{-1}.
    \end{equation}
\end{proposition}
\begin{proof}
    The proof uses repeatedly standard properties of the Kronecker product and the so-called ``vec-trick'', i.e., $\vecto(ABC)=(C^\top \otimes A)\vecto(B)$ for any matrices $A,B,C$ such that $ABC$ is well-defined.
    
    Let us start by rewriting $b_{t-1}$ as follows:
    \begin{align*}
        b_{t-1} &= \sum_{s=1}^{t-1}X_s^\top y_s = \sum_{s=1}^{t-1} (W_s^\top \otimes V)^\top \vecto(Y_s) \\
        &= \sum_{s=1}^{t-1} (W_s \otimes V^\top) \vecto(Y_s) = \sum_{s=1}^{t-1} \vecto(V^\top Y_s W_s^\top).
    \end{align*}
    Then, we also rewrite $A_t$ as follows:
    \begin{align*}
        A_t &= \Lambda_t + \sum_{s=1}^t X_s^\top X_s\\
        &= \tilde{\Lambda}\otimes (V^\top V) + \sum_{s=1}^t (W_s^\top\otimes V)^\top(W_s^\top\otimes V)\\
        &= \left(\tilde{\Lambda} + \sum_{s=1}^t W_s W_s^\top\right) \otimes (V^\top V).
    \end{align*}
    Thus, if we define $\tilde{A}_t = \tilde{\Lambda} + \sum_{s=1}^t W_s W_s^\top$, we have that $A_t^{-1}=\tilde{A}_t^{-1}\otimes (V^\top V)^{-1}$. Finally, we obtain that:
    \begin{align*}
        \vecto(\Theta_t) &= A_t^{-1}b_{t-1}\\
        &= \big(\tilde{A}_t^{-1}\otimes (V^\top V)^{-1}\big)\vecto\Big(\sum_{s=1}^{t-1} V^\top Y_s W_s^\top\Big)\\
        &= \vecto\Big( (V^\top V)^{-1} \Big(\sum_{s=1}^{t-1} V^\top Y_s W_s^\top\Big) \tilde{A}_t^{-1}\Big),
    \end{align*}
    which leads us to the desired expression.
\end{proof}
The special regularization matrix that appears in Proposition \ref{special_reg_prop} may be seen as a way of incorporating prior information (the fact that $V_t=V$) in the regularization. Indeed, for $\Lambda_t=\tilde{\Lambda}\otimes (V^\top V)$, we have
\begin{equation*}
    \norm{\vecto{\Theta}}_{\Lambda_t}^2 = \norm{V\Theta\tilde{\Lambda}^{1/2}}_F^2,
\end{equation*}
where $\tilde{\Lambda}^{1/2}$ is the positive square root of $\tilde{\Lambda}\succ 0$.

As a consequence of Proposition \ref{special_reg_prop}, an efficient implementation of \eqref{multivaw_eq} with special regularization can be proposed. Let us define: 
\begin{equation*}
    \tilde{A}_t = \tilde{\Lambda} + \sum_{s=1}^t W_s W_s^\top
    \text{ and }
    \tilde{b}_t = \sum_{s=1}^{t} Y_s W_s^\top.
\end{equation*}
Then, the complexity per time step boils down to:
\begin{enumerate}
    \item Computing $\tilde{A}_t^{-1}$. As discussed earlier this can be done recursively in a direct way or using Woodbury formula costing $O(m^3+k_t m^2)$ or $O(k_t^3+k_t^2 m + k_t m^2)$ respectively.
    \item Computing the prediction $\hat{Y}_t=V\Theta_t W_t$ by exploiting the fact that some of the matrices that appear can be computed offline. This step adds a cost of $O(ndk_t)$ or $O(n^2 k_t)$ depending on the order used for computations. 
    \item Computing $\tilde{b}_t$ recursively which finally adds a cost of $O(n k_t m)$ operations.
\end{enumerate}
To sum-up, in the setting of Proposition \ref{special_reg_prop} we reduced the complexity from a polynomial of degree six to a cubic polynomial. Finally, let us highlight that Proposition \ref{special_reg_prop} is applicable in the online multi-output regression problem (see Example \ref{Ex:framework online multiple output regression}) where $V_t=I_n$, in such a case the special regularization of Proposition \ref{special_reg_prop} writes $\Lambda_t =\tilde{\Lambda}\otimes I_n$ and generalizes the standard time-invariant regularization $\Lambda_t = \lambda I_{nm}$.

\subsection{Hierarchical Forecasting}
Since OHF corresponds to the alternative model of Section \ref{equivalent_model_appendix} with $k_t=1$, $V_t=S_t$ and $W_t=x_t$, MultiVAW-OHF has a complexity per time step that scales like $O(m^3d^3+n_t m^2d^2)$. As discussed earlier, when specialized to time-invariant regularization the Woodburry formula provides the alternative complexity: $O(n_t m^2d^2+n_t^2md+n_t^3)$. This can be further improved when implementing MetaVAW. Indeed, by exploting the closed-form expression \eqref{metavaw_close_eq} using the Woodburry formula and computing offline $S^\dagger$ one can implement MetaVAW in a $O(m^2+nm+nd)$ operations per time step. 

\section{Proofs}

\subsection{Proof of Theorem \ref{main_theorem}}\label{S:proof of main result}

    The proof of Theorem \ref{main_theorem} starts by using standard arguments that appear in the analysis of \eqref{vaw_eq} \citep[Theorem 2]{olr_gaillard_2019}. However, instead of concluding using Lemma 15 of \citet{olr_gaillard_2019} or Lemma 11.11 of \citet{prediction_cesa_2006} we need another technical result, namely Lemma 12 of \citet{log_regret_hazan_2007}, which is recalled below.
    
    \begin{lemma}[Lemma 12 of \citet{log_regret_hazan_2007}]
        \label{hazan_lemma}
        Let $A \succeq B \succ 0$ be symmetric positive definite matrices. Then, 
        \begin{equation*}
            \tr(A^{-1}(A-B)) \leq \sum_{i=1}^d\log\left(\frac{\lambda_i(A)}{\lambda_i(B)}\right).
        \end{equation*}
    \end{lemma}
    
    Let us prove Theorem \ref{main_theorem}. Let $\Lambda_0$ be an arbitrary matrix satisfying $0\prec\Lambda_0\preceq \Lambda_1$, and define $A_0 = \Lambda_0$, $b_0=0$, and for each $t=0,\dots,T$, the auxiliary function:
    \begin{equation*}
    \tilde{L}_t(\theta) = \norm{\theta}_{\Lambda_t}^2 + \sum_{s=1}^{t}\norm{X_s\theta-y_s}^2_2 =\theta^\top A_t\theta+\sum_{s=1}^{t}\norm{y_s}^2_2-2b_t^\top \theta,
    \end{equation*}
    and the vector $\tilde{\theta}_{t+1}=\argmin_{\theta\in\R^d}\tilde{L}_t(\theta)$ which is uniquely defined by the closed-form expression $\tilde{\theta}_{t+1}=A_t^{-1}b_t$. 
    Note the difference with the iterates of \eqref{multivaw_eq}, which verify ${\theta}_{t+1}=A_{t+1}^{-1}b_t$ instead (see Eq. \eqref{multivaw_close_eq}).
    Exploiting the fact that $b_t = A_t\tilde{\theta}_{t+1}$ we can now write:
    \begin{equation*}
\min_{\theta\in\R^d}\tilde{L}_t(\theta)=\tilde{L}_t(\tilde{\theta}_{t+1})=\sum_{s=1}^{t}\norm{y_s}^2_2-\tilde{\theta}_{t+1}^\top A_t \tilde{\theta}_{t+1}.
    \end{equation*}
    Now, we relate $R_T(\theta)$ to the functions $\tilde{L}_t$. We have:
    \begin{align}
         R_T(\theta) &= \sum_{t=1}^T \norm{X_t\theta_t-y_t}^2_2-\sum_{t=1}^T \norm{X_t\theta-y_t}^2_2 = \sum_{t=1}^T \norm{X_t\theta_t-y_t}^2_2 - \tilde{L}_T(\theta) + \norm{\theta}_{\Lambda_T}^2 \nonumber \\
        & \leq \sum_{t=1}^T \norm{X_t\theta_t-y_t}^2_2 - \tilde{L}_T(\tilde{\theta}_{T+1}) + \norm{\theta}_{\Lambda_T}^2 = \norm{\theta}_{\Lambda_T}^2 + \sum_{t=1}^T \Big(\norm{X_t\theta_t-y_t}^2_2 + \tilde{L}_{t-1}(\tilde{\theta}_{t}) - \tilde{L}_{t}(\tilde{\theta}_{t+1}) \Big). \label{first_tmp_regret_bound_eq}
    \end{align}
    Let us now bound the terms inside the sum of the last inequality. For each $t\in[T]$, we have by definition of $A_t$ and $b_t$ that $X_t^\top X_t= A_t-A_{t-1}+\Lambda_{t-1}-\Lambda_t$ and $X_t^\top y_t = b_t-b_{t-1}= A_t(\tilde{\theta}_{t+1}-\theta_t)$, thus we can write:
    \begin{align*}
         &\norm{X_t\theta_t-y_t}^2_2 + \tilde{L}_{t-1}(\tilde{\theta}_{t}) - \tilde{L}_{t}(\tilde{\theta}_{t+1}) \\
        &\quad =\theta_t^\top X_t^\top X_t \theta_t - 2y_t^\top X_t\theta_t -\tilde{\theta}_{t}^\top A_{t-1} \tilde{\theta}_{t} + \tilde{\theta}_{t+1}^\top A_{t}\tilde{\theta}_{t+1} \\
        &\quad = \theta_t^\top(A_t-A_{t-1} + \Lambda_{t-1}-\Lambda_t)\theta_t-2(X_t^\top y_t)^\top \theta_t -\tilde{\theta}_{t}^\top A_{t-1} \tilde{\theta}_{t} + \tilde{\theta}_{t+1}^\top A_{t}\tilde{\theta}_{t+1}\\
        &\quad = \theta_t^\top(A_t-A_{t-1})\theta_t - \theta_t^\top(\Lambda_t-\Lambda_{t-1})\theta_t-2(A_t(\tilde{\theta}_{t+1}-\theta_t))^\top \theta_t  -\tilde{\theta}_{t}^\top A_{t-1} \tilde{\theta}_{t} + \tilde{\theta}_{t+1}^\top A_{t}\tilde{\theta}_{t+1}.
    \end{align*}
     Then, we reorder the terms and rewrite them using $A_t\theta_t=b_{t-1}=A_{t-1}\tilde{\theta}_t$, which yields:
    \begin{align*}
        & \norm{X_t\theta_t-y_t}^2_2 + \tilde{L}_{t-1}(\tilde{\theta}_{t}) - \tilde{L}_{t}(\tilde{\theta}_{t+1}) \\
        &\quad = (\tilde{\theta}_{t+1}^\top A_{t}\tilde{\theta}_{t+1} + \theta_{t}^\top A_{t}\theta_{t} - 2 \tilde{\theta}_{t+1}^\top A_t \theta_{t} ) - (\tilde{\theta}_{t}^\top A_{t-1}\tilde{\theta}_{t} + \theta_{t}^\top A_{t-1}\theta_{t} - 2 \theta_{t}^\top (A_t \theta_{t})) - \theta_t^\top(\Lambda_t-\Lambda_{t-1})\theta_t \\
        &\quad = (\tilde{\theta}_{t+1}^\top A_{t}\tilde{\theta}_{t+1} + \theta_{t}^\top A_{t}\theta_{t} - 2 \tilde{\theta}_{t+1}^\top A_t \theta_{t} ) - (\tilde{\theta}_{t}^\top A_{t-1}\tilde{\theta}_{t} + \theta_{t}^\top A_{t-1}\theta_{t} - 2 \theta_{t}^\top (A_{t-1} \tilde{\theta}_{t})) - \theta_t^\top(\Lambda_t-\Lambda_{t-1})\theta_t \\
        &\quad = (\tilde{\theta}_{t+1}-\theta_t)^\top A_t (\tilde{\theta}_{t+1}-\theta_t) - (\tilde{\theta}_{t}-\theta_t)^\top A_{t-1} (\tilde{\theta}_{t}-\theta_t) - \theta_t^\top(\Lambda_t-\Lambda_{t-1})\theta_t.
    \end{align*}
    
    Since $A_{t-1}\succeq 0$, $\Lambda_t-\Lambda_{t-1}\succeq 0$ and $\tilde{\theta}_{t+1}-\theta_t = A_t^{-1}(b_t-b_{t-1})=A_t^{-1}X_t^\top y_t$, we obtain:
    \begin{equation*}
        \norm{X_t\theta_t-y_t}^2_2 + \tilde{L}_{t-1}(\tilde{\theta}_{t}) - \tilde{L}_{t}(\tilde{\theta}_{t+1})
        \leq (\tilde{\theta}_{t+1}-\theta_t)^\top A_t (\tilde{\theta}_{t+1}-\theta_t) 
        = y_t^\top X_t A_t^{-1} X_t^\top y_t.
    \end{equation*}
    Summing these inequalities, combining them with \eqref{first_tmp_regret_bound_eq}, and observing that $X_t A_t^{-1} X_t^\top\succeq 0$ leads to:
    \begin{equation}
    R_T(\theta) \leq \norm{\theta}_{\Lambda_T}^2 + \sum_{t=1}^T y_t^\top X_t A_t^{-1} X_t^\top y_t \leq \norm{\theta}_{\Lambda_T}^2 + \sum_{t=1}^T y_t^\top \lambda_1(X_t A_t^{-1} X_t^\top) y_t
    \leq \norm{\theta}_{\Lambda_T}^2 + \bar{y}^2 \sum_{t=1}^T \tr(X_t A_t^{-1} X_t^\top). \label{last_tmp_regret_bound_eq}
    \end{equation}

    On the other hand, applying Lemma \ref{hazan_lemma} to the matrices $B=A_{t-1}-\Lambda_{t-1}+\Lambda_t$, $A=A_t=B+X_t^\top X_t$ which satisfy $A\succeq B \succ 0$ leads us to:
    \begin{equation*}
        \tr(A_t^{-1} X_t^\top X_t) \leq \sum_{i=1}^d\log\left(\frac{\lambda_i(A_{t})}{\lambda_i(A_{t-1}-\Lambda_{t-1}+\Lambda_t)}\right),
    \end{equation*}
    for all $t\in[T]$. Furthermore, since we have both $A_{t-1}\succeq 0$ and $\Lambda_t-\Lambda_{t-1}\succeq 0$, a standard corollary of Weyl's inequality ensures that for all $i\in[d]$ we have: $\lambda_i(A_{t-1}) \leq \lambda_i(A_{t-1}-\Lambda_{t-1}+\Lambda_t)$.

    The final step of this proof consist of summing the resulting inequalities and telescoping:
    \begin{equation*}
    \sum_{t=1}^T \tr(X_tA_t^{-1}X_t^\top) = \sum_{t=1}^T \tr(A_t^{-1}X_t^\top X_t)  \leq \sum_{t=1}^T \sum_{i=1}^d \log\left(\frac{\lambda_i(A_t)}{\lambda_i(A_{t-1})}\right) = \sum_{i=1}^d \log\left(\frac{\lambda_i(A_T)}{\lambda_i(\Lambda_0)}\right).
    \end{equation*}
    Combining this with $\eqref{last_tmp_regret_bound_eq}$ and choosing $\Lambda_0=\Lambda_1$ concludes the proof.

\subsection{Proof of Corollary \ref{standard_reg_cor}}
Let us fix $\Lambda_t=\lambda I_d$ and apply Jensen's inequality to the regret bound of Theorem \ref{main_theorem}, this leads us to:
\begin{equation*}
    R_T(\theta)\leq \lambda \norm{\theta}_2^2 + d \bar{y}^2\log\left(\sum_{i=1}^d\frac{\lambda_i(A_T)}{d\lambda}\right).
\end{equation*}
We can further simplify this bound by observing that:
\begin{equation*}
    \sum_{i=1}^d\lambda_i(A_T) = \tr\Big(\lambda I_d + \sum_{t=1}^T X_t^\top X_t\Big)= \lambda d + \sum_{t=1}^T \norm{X_t}_F^2.
\end{equation*}
We conclude by using $\norm{X_t}_F^2\leq \bar{X}^2$.
\subsection{Proof of Proposition \ref{metavaw_close_prop}}
    Let us denote by $\tilde{y}_{t,i}$ the base forecasts of MetaVAW. Since they are generated by \eqref{vaw_eq} they can be written as $\tilde{y}_{t,i} = \theta_{t,i}^\top x_t$, where:
    \begin{equation*}
        \theta_{t,i} = \left(\lambda I_m + \sum_{s=1}^t x_s x_s^\top\right)^{-1} \sum_{s=1}^{t-1}x_s y_{s,i}.    
    \end{equation*}
    Thus, we have:
    \begin{equation*}
        \tilde{y}_{t,i} = \left(\sum_{s=1}^{t-1}y_{s,i} x_s^\top \right) \left(\lambda I_m + \sum_{s=1}^t x_s x_s^\top\right)^{-1}x_t.
    \end{equation*}
    Stacking the base forecasts into a vector $\tilde{y}_t\in\R^n$, we obtain that:
    \begin{equation*}
        \tilde{y}_t = \left(\sum_{s=1}^{t-1}y_s x_s^\top \right) \left(\lambda I_m + \sum_{s=1}^t x_s x_s^\top\right)^{-1}x_t.
    \end{equation*}
    Since the last step of MetaVAW consist of projecting $\tilde{y}_t$ into the set of coherent signals $\image S$, we just derived expression \eqref{metavaw_close_eq}.

\subsection{Proof of Proposition \ref{meta_is_multi_prop}}
The OHF problem with time-invariant constraints corresponds to the setup of Proposition \ref{special_reg_prop} available in Appendix \ref{S:computational cost matrix estimation} where $V=S$ and $W_t=x_t$. Therefore, Proposition \ref{special_reg_prop} provides the following closed-form expression for $\Theta_t$, the parameters of MultiVAW-OHF with regularization $\Lambda_t =\lambda I_m\otimes (S^\top S)$, 
\begin{equation*}
    \Theta_t = (S^\top S)^{-1}S^\top \left(\sum_{s=1}^{t-1} y_s x_s^\top \right)\left(\lambda I_m + \sum_{s=1}^t x_s x_s^\top\right)^{-1}.
\end{equation*}
Additionally, notice that since $S$ is injective we have $S^\dagger = (S^\top S)^{-1}S^\top$. Therefore, the expression of the predictions of MultiVAW-OHF with regularization $\Lambda_t =\lambda I_m\otimes (S^\top S)$ which write $\hat{y}_t=S\Theta_t x_t$ coincide with those of MetaVAW \eqref{metavaw_close_eq}.

\subsection{Proof of Corollary \ref{ohf_bounds_cor}}
\begin{enumerate}
    \item
    The bound claimed in the case where $\Lambda_t =\lambda I_{dm}$ is a direct consequence of Corollary \ref{standard_reg_cor} where we simply observed that $\norm{X_t}_F^2=\norm{x_t^\top \otimes S}_F^2=\norm{x_t}_2^2\norm{S_t}_F^2 \leq \bar{x}^2\bar{S}^2$ .
    \item According to Proposition \ref{meta_is_multi_prop}, MetaVAW is exactly MetaVAW-OHF with $\Lambda_t =\lambda I_m \otimes(S^\top S)$.
    The regret bound we obtained in Theorem \ref{main_theorem} gives us:
\begin{equation*}
    \label{main_matrix_bound_eq}
    R_T(\Theta) \leq 
    \norm{\vecto\Theta}_{\Lambda_T}^2
        +\bar{y}^2\sum_{i=1}^{dm} \log\left(\frac{\lambda_i(A_T)}{\lambda_i(\Lambda_1)}\right).
\end{equation*}
    Applying Jensen's inequality leads to:
    \begin{equation*}
    R_T(\Theta) \leq 
        \norm{\vecto\Theta}_{\Lambda_T}^2
        +dm\bar{y}^2\log\left(\sum_{i=1}^{dm} \frac{\lambda_i(A_T)}{dm\lambda_i(\Lambda_1)}\right).
    \end{equation*}
    Also, recall that $\norm{\vecto(\Theta)}_{\lambda I_m\otimes (S^\top S)}^2=\lambda\norm{S\Theta}_F^2$, thus, using $\lambda_i(\Lambda_1)\geq\lambda_{dm}(\Lambda_1)=\lambda\lambda_d(S^\top S)$ for all $i\in[dm]$ and $A_T = \tilde{A}_T\otimes (S^\top S)$ where $\tilde{A}_T = \lambda I_m+\sum_{t=1}^T x_t x_t^\top$, we can derive from the previous inequality the following bound:
    \begin{equation*}
        R_T(\Theta) \leq \lambda\norm{S\Theta}_F^2 + dm \bar{y}^2\log\left(\frac{\tr(\tilde{A}_t\otimes (S^\top S))}{dm\lambda\lambda_d(S^\top S)}\right).
    \end{equation*}
    Finally, using $\tr(\tilde{A}_t\otimes (S^\top S))=\tr(\tilde{A}_t)\tr(S^\top S)$, the linearity of the trace and $\norm{x_t}_2^2\leq \bar{x}^2$ we obtain \eqref{metavaw_bound_eq}.
\end{enumerate}

\end{document}